 \documentclass[pmlr,twocolumn,10pt]{jmlr} 

\usepackage{booktabs}
\usepackage{siunitx}

\usepackage{color,graphicx}
\usepackage{amsmath,amssymb,amsopn,bm,mathtools}
\usepackage{url}
\usepackage{multirow}
\usepackage{todonotes}
\usepackage[utf8]{inputenc}
\usepackage{dirtytalk}

\theorembodyfont{\upshape}
\theoremheaderfont{\scshape}
\theorempostheader{:}
\theoremsep{\newline}

\jmlrvolume{LEAVE UNSET}
\jmlryear{2023}
\jmlrsubmitted{LEAVE UNSET}
\jmlrpublished{LEAVE UNSET}
\jmlrworkshop{Machine Learning for Health (ML4H) 2023} 

\title[Validity problems in clinical machine learning]{Validity problems in clinical machine learning by indirect data labeling using consensus definitions\titletag{\thanks{This article is an extended version of Chapter 2.4.3 of \citet{RiezlerHagmann:21}.}}}

 \author{%
  \Name{Michael Hagmann} \Email{hagmann@cl.uni-heidelberg.de}\\
  \Name{Shigehiko Schamoni} \Email{schamoni@uni-heidelberg.de}\\
  \Name{Stefan Riezler} \Email{riezler@cl.uni-heidelberg.de}\\
  \addr Department of Computational Linguistics \\ \& Interdisciplinary Center for Scientific Computing (IWR) \\ Heidelberg University, Germany
 }

\begin{document}

\maketitle

\begin{abstract}
We demonstrate a validity problem of machine learning in the vital application area of disease diagnosis in medicine. It arises when target labels in training data are determined by an indirect measurement, and the fundamental measurements needed to determine this indirect measurement are included in the input data representation. Machine learning models trained on this data will learn nothing else but to exactly reconstruct the known target definition. Such models show perfect performance on similarly constructed test data but will fail catastrophically on real-world examples where the defining fundamental measurements are not or only incompletely available. We present a general procedure allowing identification of problematic datasets and black-box machine learning models trained on them, and exemplify our detection procedure on the task of early prediction of sepsis.
\end{abstract}
\begin{keywords}
validity, consensus definitions, early disease diagnosis
\end{keywords}

\section{Introduction}

A machine learning research project is naturally divided into the tasks of data curation --- careful preparation of training, development and test data --- and of machine learning itself, where the practitioner only needs to focus on improving the current state-of-the-art on the benchmark data, often being agnostic about the compilation process of the data. 
This division of labor incurs the risk of obtaining problematic models if the data curator determines the target label by an indirect measurement,
and the fundamental measurements used for its determination are part of the input feature representation for machine learning.
Under this setup, the obtained model will be nothing else than a nearly identical reconstruction of the applied target definition.
This constitutes a validity problem that leads to serious practical problems when the model is applied to real-world data. Particularly, the model fails on data where the fundamental measurements of the target definition are not or only incompletely available, because it hasn't learned alternative predictive patterns relying on other features. Furthermore, usages of machine learning to discover new explanatory predictive patterns will fail for exactly the same reason. 



Machine learning research in health care is exposed to this risk if labels are created by indirect measurements given by consensus definitions of medical diseases. For example, 
the Sepsis-3 definition \citep{SingerETAL:16,SeymourETAL:16} defines sepsis by the presence of a verified or suspected infection in combination with an organ function deterioration. For patients without or unknown pre-conditions, this deterioration is in practice operationalized by a total SOFA score of at least 2 points. The SOFA scoring system itself  \citep{VincentETAL:96} is defined for 6 organ systems (e.g., liver or kidney) whose scores are defined by thresholds on measurable physiological quantities like creatinine, bilirubin, urine output, mean arterial pressure, etc.  As shown in a recent overview \citep{MoorETAL:21} that examined 22 studies on machine learning approaches for sepsis prognosis, with the exception of one, all studies define ground-truth sepsis labels using rules such as the Sepsis-3 consensus definition, and use datasets such as MIMIC \citep{JohnsonETAL:16} for training and testing. Similarly, a recent challenge on early prediction of sepsis from clinical data assigned labels in accordance with the Sepsis-3 consensus definition \citep{ReynaETAL:19}. The approaches described in \cite{MoorETAL:21} and \cite{ReynaETAL:19} include the fundamental physiological measurements involved in the consensus definition of sepsis as input features for machine learning, which makes them  \emph{defining features} in the indirect measurement of the consensus definition. 

Given the negative impact of defining features, we need a procedure to identify them in training data.
Starting from a set of candidate features,
e.g, identified by the absolute magnitude of their bivariate correlation with the labels, 
we present the following two-step detection procedure, based on interpretable \emph{Generalized Additive Models (GAM)} :
\begin{enumerate}
\item In a first step, we 
search for the GAM with the \emph{best data fit} and \emph{smallest complexity} that can be built using only candidate features. 
\item In a second step, we confirm that the features constituting the model identified in the first step are indeed \emph{defining} by checking if the contribution of all other candidate features to the prediction is \emph{nullified} in a model that includes the defining features as inputs. 
\end{enumerate}
 We exemplify the problem on the consensus definitions involved in SOFA score and more complex examples like the Sepsis-3 consensus definition. The description of the experiments will rely on the informally defined steps of the above detection procedure. Formal details are given in the Appendix. Code and data to reproduce this work are available at \url{https://github.com/StatNLP/ml4h_validity_problems}.



\section{Experiments}
\label{sec:exps}

\subsection*{Data}
The input data used in our experiments consist of several clinical measurements for $620$ intensive care patients from the surgical intensive care unit of the University Medical Centre Mannheim, Germany (see \cite{LindnerETAL:22,SchamoniETAL:22} for a detailed description). We split the data into train ($323{,}404$ measurement points) and test ($80{,}671$ measurement points) partitions. Furthermore, we label the data according to the Sepsis-3 consensus definition \citep{SingerETAL:16,SeymourETAL:16} that combines the criteria of SOFA and infection.

\begin{table}[t!]
        \centering
        \caption{Definition of liver SOFA score. Serum bilirubin levels are compared to specific thresholds listed in $mg/dl$ to determine the SOFA$_{liver}$ score.}
        \label{fig:definition_lsofa}
        \includegraphics[width=0.6\columnwidth]{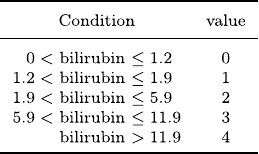}
    \end{table}
\begin{table}[h!]
      \centering
        \caption{Step 1 of detection of defining features for liver SOFA: Top 5 GAMs for liver SOFA, ranked by data fit(↓) and complexity(↑).}
        \label{fig:search_lsofa}
        \includegraphics[width=\columnwidth]{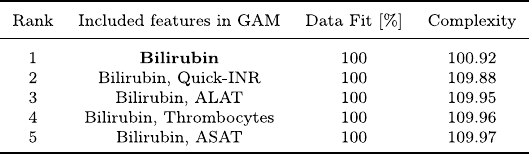}
    \end{table}

\subsection*{Detecting defining features in datasets for liver SOFA scores}
In the following example, we assume that  $D = \{(\mathbf{x}^n,{y}^n)\}_{n=1}^N$ is a dataset of input features $\mathbf{x} = (x_1, x_2, \ldots, x_p)$ and gold standard labels $y$ of liver SOFA scores. We apply the two-step detection procedure described above to this data. That means, we train a GAM on this dataset, and use the statistical metrics of data fit, model complexity, and nullification to expose fundamental measurements in $\mathbf{x}$ as defining features. This is achieved by showing that they allow an exact reconstruction of the target functional definition $y$, while nullifying the contribution of all other features. 

\begin{figure*}[t!]
        \centering
        \includegraphics[width=0.6\textwidth]{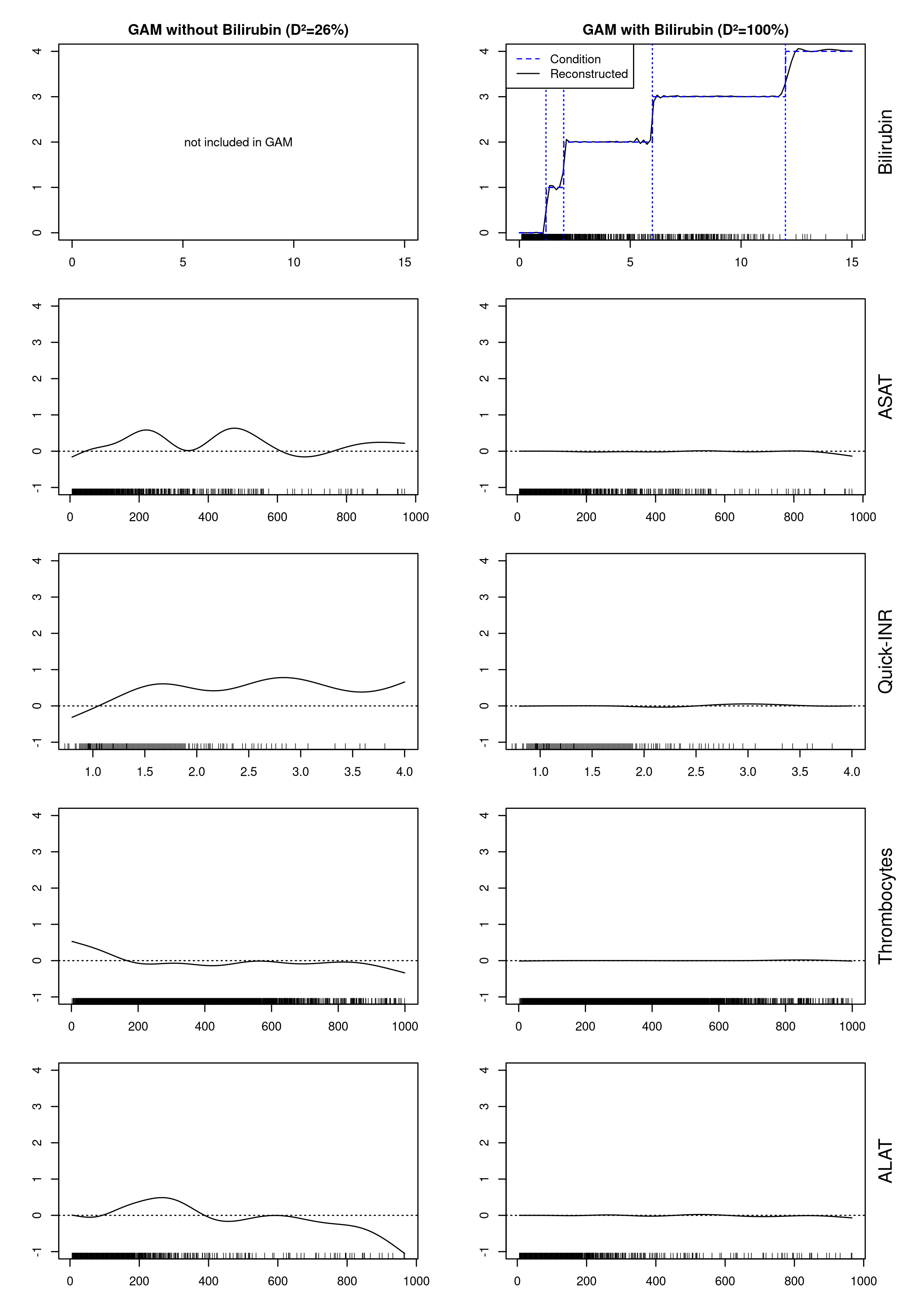}
        \caption{Step 2 of detection of defining features for liver SOFA: Feature shapes and data fit $D^2$ of GAMs for liver SOFA data without (left column) and with bilirubin (right column) as input feature, showing perfect reconstruction of the target definition and nullfication of non-defining features.}
        \label{fig:lsofa_nullification}
\vspace{-1em}
\end{figure*}

For ease of presentation, we choose a subset of the original $43$ features used in \cite{SchamoniETAL:22}. In the liver SOFA example we consider the clinical measurements of bilirubin, aspartate aminotransferase (ASAT), a quick international normalized ratio of the protein prothrombin (quick-INR), alanin aminotransferase (ALAT), and thrombocytes.
\footnote{The features were selected based on the magnitude of Pearson correlation with the target.}
As shown in 
Table \ref{fig:definition_lsofa}, 
bilirubin is a fundamental measurement for the indirect measurement of liver SOFA.

 Step 1 of the detection procedure trains a GAM on the five candidate features, yielding the result shown in Table \ref{fig:search_lsofa}. We see that any model including bilirubin as feature explains the data with a data fit $D^2 = 100\%$. The model that uses  bilirubin as only feature has the smallest complexity. Step 2 of the detection procedure is depicted in Figure \ref{fig:lsofa_nullification}. It shows the feature shapes of GAMs fitted to subsets of the training data with (right column) and without bilirubin (left column). The plots show that the GAM trained without access to bilirubin uses all other available features (rows 2--5 in left panel), whereas their contribution to the prediction is nullified in the GAM trained on data including bilirubin (rows 2--5 in right panel). The comparison of the data fit metric $D^2$ for both GAMs illustrates that the inclusion of bilirubin in the training data boosts the data fit from $26\%$ to $100\%$. This means that the GAM trained on data including bilirubin is able to perfectly replicate the liver SOFA scores for all test inputs. Additionally, the first panel of the right column demonstrates that a GAM is able to nearly perfectly reconstruct the liver SOFA definition from training data including the defining feature bilirubin: The steps of the plateaus in the graph correspond exactly to the thresholds listed in Table \ref{fig:definition_lsofa}. 
This shows that our detection procedure is able to successfully identify bilirubin as a defining feature for the indirect measurement of liver SOFA, out of a candidate set of fundamental features with a similarly high bivariate correlation to the target label. 

\subsection*{Detecting defining features in datasets for kidney SOFA scores}

\begin{table}[t!]
        \centering
        \caption{Definition of kidney SOFA score. Serum creatinine levels and urine output are compared to thresholds listed in $mg/dl$ and $ml/day$, respectively, to determine their component value. The final score is the maximum of both: SOFA$_{kidney}$ $= \mathrm{max}(\text{value}_{crea},\text{value}_{urine})$.}
        \label{fig:definition_ksofa}
        \includegraphics{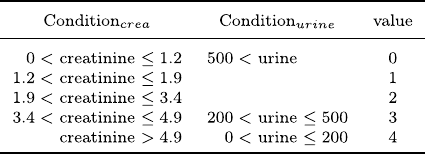}
    \end{table}

\begin{table}[t!]
        \centering
        \caption{Step 1 of two-step detection of defining features for kidney SOFA: Top 5 GAMs for kidney SOFA, ranked by data fit($\downarrow$) and complexity ($\uparrow$).}
        \label{fig:search_ksofa}
        \includegraphics[width=\columnwidth]{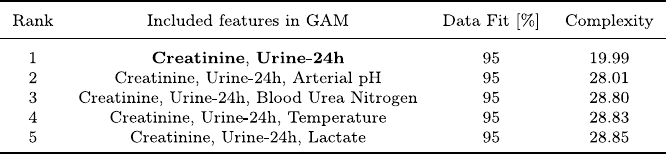}
    \end{table}

\begin{figure}[t!]
        \centering
   \includegraphics[width=\columnwidth]{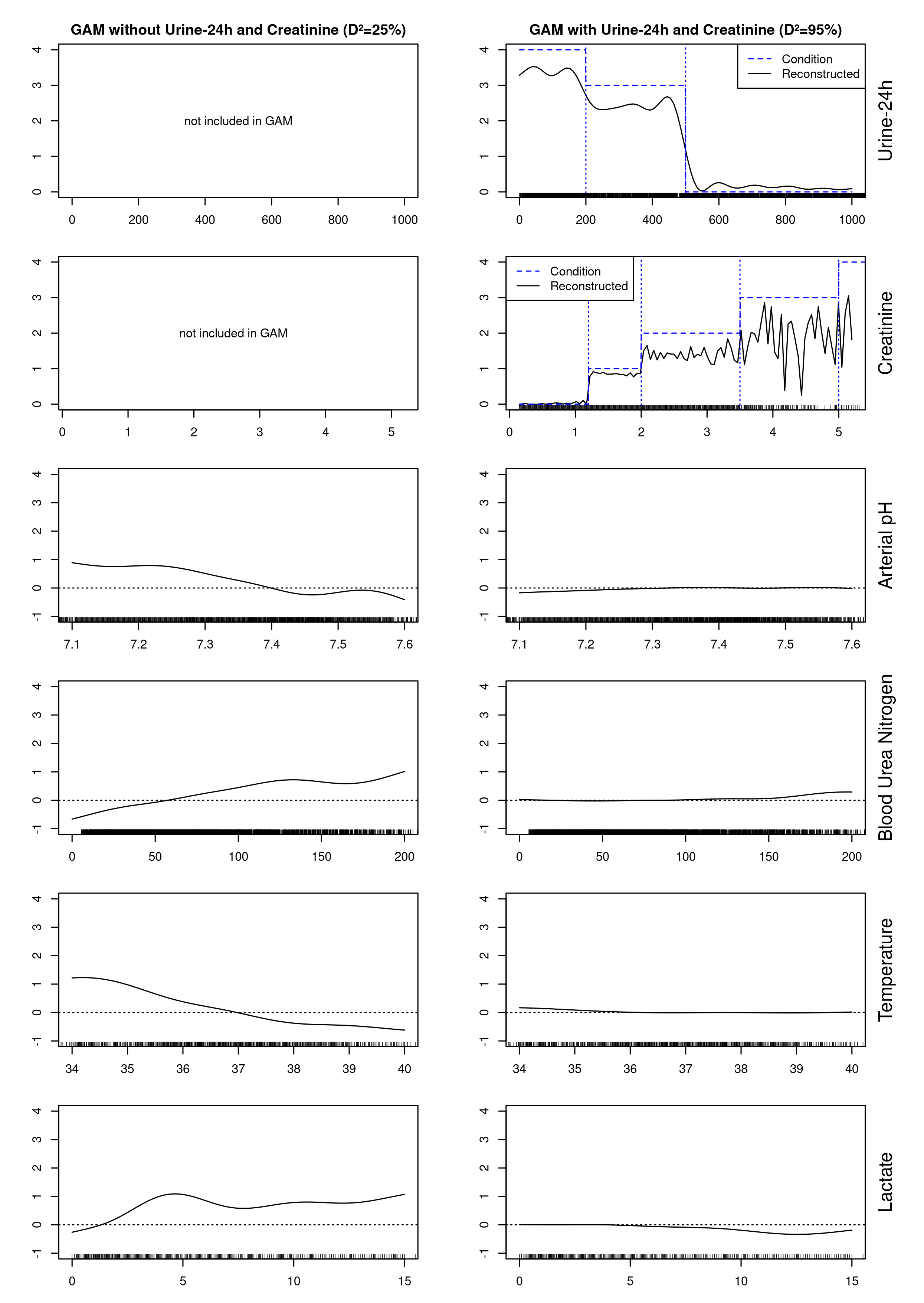}
\caption{Step 2 of two-step detection of defining features for kidney SOFA: Feature shapes and data fit $D^2$ of GAMs trained on kidney SOFA data without (left col.) and with creatinine and urine-24h (right col.) as input features, showing reconstruction of target definition and nullification of non-defining features.
}
\label{fig:ksofa_circularity}
\vspace{-1.5em}
\end{figure}

As shown in Table \ref{fig:definition_ksofa}, the kidney SOFA score is defined as a maximum of two step functions. This bivariate function cannot be expressed as sum of smooth functions using spline-based GAMs. Additionally, we face the problem that creatinine and urine-24h output are significantly correlated ($r = -.28$, $p<.0001$). However, as we will see, a GAM-based analysis of the univariate feature shapes for urine-24h and creatinine still yields some useful insights. Let us first apply step 1 of the detection procedure for defining features to data with automatically assigned kidney SOFA labels. Table \ref{fig:search_ksofa} displays the ranking of models using the 6 candidate features, yielding a data fit $D^2 = 95\%$ for any model including creatinine and urine-24h as features, where the model that includes only these two features has the smallest complexity.

Figure \ref{fig:ksofa_circularity} depicts the feature shapes of a model using all features in the dataset (right column) and a model that was trained  without creatinine and urine-24h (left column). The $D^2$ value of the model on the right reaches $95\%$, compared to $25\%$ for the model without defining features. The bottom four plots in the right column show that the contribution of any feature in the model without creatinine and urine-24h (left column) is nullified by inclusion of urine-24h and creatinine as features. Furthermore, the top two plots in the right column show that the target definition of kidney SOFA given in Table \ref{fig:definition_ksofa} can be satisfactorily reconstructed, despite the fact that the kidney SOFA is a composition of two step functions. By combining the conditions of data fit and nullification, our detection procedure was able to successfully identify the fundamental measurements of creatinine and urine-24h output as defining features for the indirect measurement of kidney SOFA, compared to other fundamental features with a similarly high bivariate correlation to the target label. 

\subsection*{Detecting defining features for Sepsis-3 consensus definition}

\begin{table}[t!]
        \centering
        \caption{Operationalization of Sepsis-3 according to \cite{SingerETAL:16}. A patient has sepsis if the maximum SOFA score in the last 24h is greater or equal to 2 in combination with a suspected or verified infection; expressed here as a product of the two conditions: Sepsis$_{Sepsis\text{-}3} =  \text{Condition}_{Infection} \times \text{Condition}_{SOFA}$.}
        \label{fig:definition_sepsis}
        \includegraphics[width=\columnwidth]{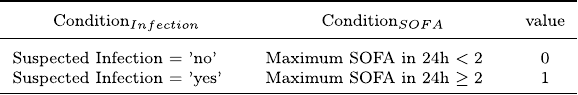}
\end{table}

\begin{table}[t!]
        \centering
        \caption{Step 1 of two-step detection of defining features for Sepsis-3: Top 6 GAMs for Sepsis-3, ranked by data fit($\downarrow$) and complexity ($\uparrow$).}
        \label{fig:search_spesis3}
        \includegraphics[width=\columnwidth]{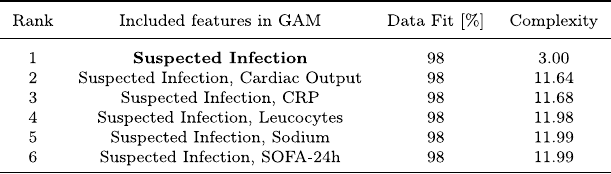}
    \end{table}
    
\begin{figure}[t!]
        \centering
   \includegraphics[width=\columnwidth]{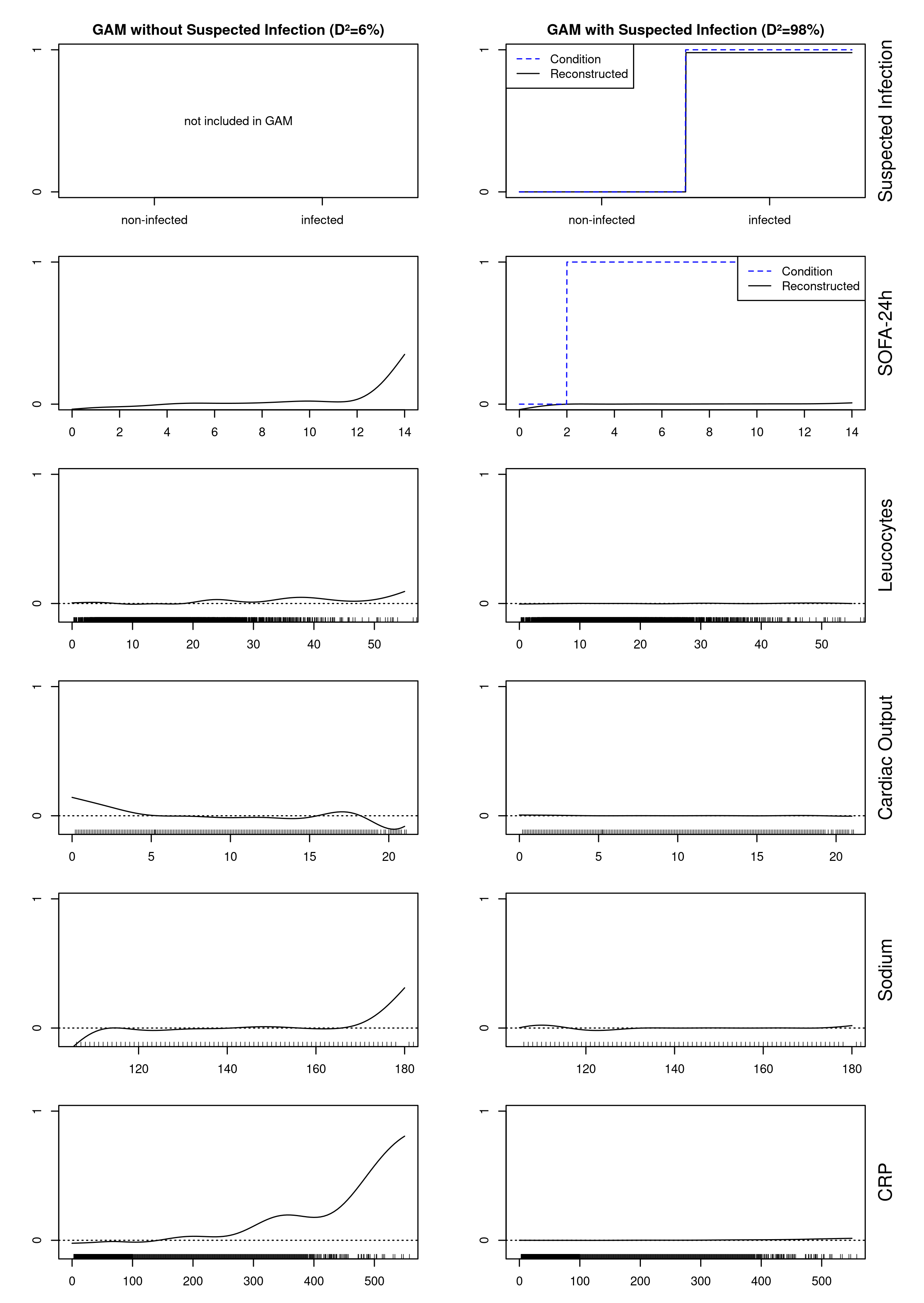}
\caption{Step 2 of two-step detection of defining features for Sepsis-3: Feature shapes and data fit $D^2$ of GAMs trained on Sepsis-3 data without (left col.) and with suspected infection (right col.) as input features, showing satisfactory reconstruction of the target definition and nullification of non-defining features. 
}
\label{fig:sepsis3_circularity}
\vspace{-1.5em}
\end{figure}

SOFA-definitions as used in the previous section are the basis for further experiments to predict sepsis according to the Sepsis-3 consensus definition \citep{SingerETAL:16,SeymourETAL:16}. Sepsis according to Sepsis-3 is present if the following two conditions are met: 1) The patient suffers from deterioration of organ functions; 
2) The patient has a suspected or verified infection.
According to \cite{SingerETAL:16}, a SOFA score of 2 points or more indicates an organ dysfunction for patients without any pre-conditions. 
A suspected or verified infection is operationalized by defining an infection window by the first dose of antibiotics and the first draw of blood culture samples (see eAppendix A of \cite{SeymourETAL:16}).\footnote{The start of the window in combination with the SOFA criterion is then the onset of sepsis. The length of the sepsis episode is determined by the infection window. Subsequent administered antibiotics and additional draws of blood samples can thus extend a sepsis episode.}
The complete definition and the conditions we use are compiled in Table \ref{fig:definition_sepsis}.

The candidate features are leukocytes, cardiac output, sodium, and C-reactive protein (CRP), suspected infection, and maximum SOFA score in 24 hours (SOFA-24h).
Step 1 of the detection procedure produces the results shown in Table \ref{fig:search_spesis3}: We see that all models including suspected infection as feature explain the data with a data fit of $D^2 = 98\%$ where the model that includes only this feature has the smallest complexity. Suspected infection is thus marked as potentially defining in step 1 of the detection procedure. 
Step 2 of the detection procedure is shown in Figure \ref{fig:sepsis3_circularity}. The left column displays the feature shapes of the features of SOFA-24h, leukocytes, cardiac output, sodium, and CRP in a GAM that does not use suspected infection as input feature. All features show a clear contribution to the prediction of sepsis. Including suspected infection as input feature to the GAM yields the feature shapes shown in the right column: The contributions of all features except suspected infection are nullified. Comparing the data fit metric $D^2$ for both GAMs shows that the inclusion of suspected infection as input feature boosts the data fit from $6\%$ to $98\%$. This means that a GAM trained on data including suspected infection is able to perfectly replicate the Sepsis-3 status of each test input. We can thus identify suspected infection as defining feature for the indirect measurement of Sepsis-3, compared to other fundamental features with the highest bivariate correlation to the target label. 

Surprisingly, SOFA-24h is not identified as defining feature, but instead is nullified. This can be explained by particularities of the dataset used in this experiment \citep{LindnerETAL:22,SchamoniETAL:22}. In this dataset, only $522$ out of $28,122$ infection observations had a SOFA-24h measurement of less than $2$. This implies that the probability of having a positive Sepsis-3 label is $0.9814$ if the patient has a suspected or verified infection. This makes it possible to reduce the conditions in the Sepsis-3 definition effectively to testing for infection, and is the reason why our method only unveiled suspected infection as the sole defining feature.

\subsection*{Detecting defining features of black-box machine learning models}

The proposed procedure can also be used to detect whether a state-of-the-art machine learning model such as a deep neural network inherits a validity problem present in the training data. This is possible without access to the actual training data or the parameters of the model, which is thus treated as a black-box model. Since a trained model is a function that maps input features to labels, we can apply the detection procedure to a dataset $T = \{(\mathbf{x}^m,\hat{y}^m)\}_{m=1}^M$ consisting of test inputs $\mathbf{x}$ and labels $\hat{y}$ predicted by the black-box model. That means, we apply the same procedure as above to a dataset of inputs and labels, with the only difference that the labels are predictions of the black-box model. 
After training a GAM on test inputs and predicted labels, we use the statistical metrics of data fit, model complexity, and nullification to detect if the trained black-box model relies on training data that was labeled by indirect measurements. 

\begin{table}[t!]
        \centering
        \caption{Step 1 of two-step detection of defining features for black-box predictions: Top 5 GAMs for black-box predictions, trained with (upper) and without (lower) bilirubin feature, ranked by data fit($\downarrow$) and complexity ($\uparrow$).} 
        \label{fig:search_bbm}
        \includegraphics[width=\columnwidth]{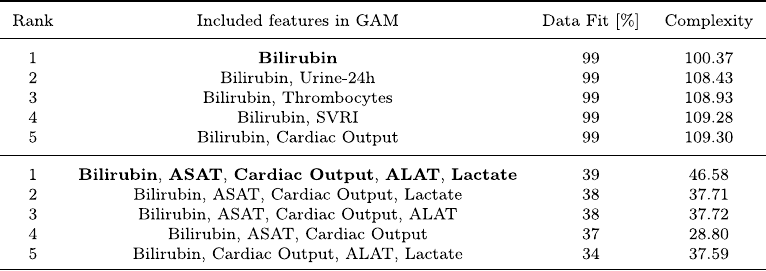}
    \end{table}

\begin{figure}[t!]
        \centering
   \includegraphics[width=\columnwidth]{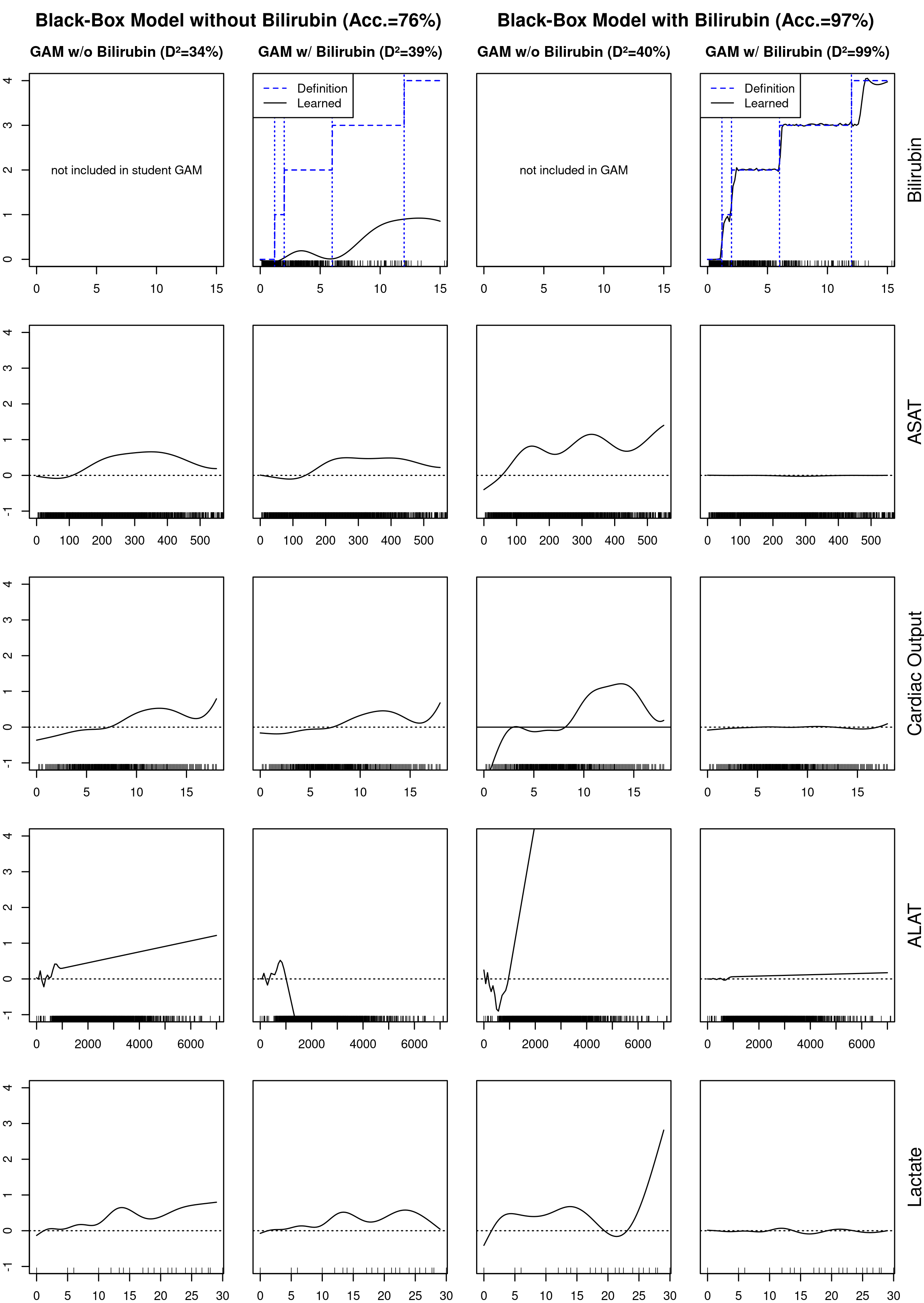}
\caption{Step 2 of two-step detection of defining features for black-box predictions: Feature shapes of GAMs trained on predictions of black-box models without (cols. 1 and 2) and with (cols. 3 and 4) access to bilirubin during training. on of target function and nullification 
shown in rightmost column.
}
\label{fig:distill_circularity_lsofa}
\vspace{-1.5em}
\end{figure}

We exemplify the detection procedure on the example of liver SOFA prediction. 
Let us start with a neural network
that has access to all $43$ original features, including bilirubin. A first noteworthy observation is that the accuracy of this neural network is $97\%$ on the test data, and $99.7\%$ on the training data.  This gap is exceptionally low, demanding a further investigation by a detection procedure for defining features.
For ease of presentation of our detection procedure, we again limit the original $43$ features to the five features that are most highly correlated with the black-box model predictions. For liver SOFA these are bilirubin, thrombocytes, cardiac output, systematic vascular resistance index (SVRI), and urine output in the previous $24$ hours (urine-24h). 
Step 1 of the detection procedure consists of training a GAM on predictions of the neural network. As shown in the upper half of Table \ref{fig:search_bbm}, the GAM with lowest complexity and highest data fit is the model using bilirubin as sole feature. This result is similar to the outcome of step 1 displayed in Table \ref{fig:search_lsofa}. The result of step 2 of our detection procedure is depicted in the fourth column of Figure \ref{fig:distill_circularity_lsofa}. The topmost panel in the fourth column shows that the predictions of the black-box model can be reproduced nearly perfectly ($D^2 = 99\%$), with a feature shape for bilirubin that is very similar to the liver SOFA definition given in Table \ref{fig:definition_ksofa}. Without access to bilirubin, the GAM in the third column of Figure \ref{fig:distill_circularity_lsofa} is shown to be only capable to poorly reproduce the black-box model's predictions ($D^2=40\%$). A comparison of the plots in the third and fourth column illustrates that the non-negligible contribution of the features in the GAM without access to bilirubin are nullified if bilirubin is included in the GAM.  We can therefore conclude that the black box model must have had access to the defining feature of bilirubin in its training data, inheriting the validity problem. 

The first and second column of Figure \ref{fig:distill_circularity_lsofa} compare this result to an application of the detection procedure to a black-box model that accesses all features except bilirubin in its training data. The test accuracy of this model drops to $76\%$, and a GAM trained on its predictions explains the data with a data fit of $D^2 = 39\%$ if bilirubin is included in the GAM, and with $D^2 = 34\%$ if bilirubin is excluded. However, the feature shapes are similar and non-null in both types of GAMs.

We furthermore conduct an ablation experiment where we deprive the neural network trained on data including bilirubin of this very feature at test time.\footnote{This is achieved by setting the feature value to $0$ (which is the mean value of the standardized input features) for all test instances, and evaluate accuracy of the new predictions.} This neural network predicts SOFA scores of $0$ in $99.7\%$ of all test predictions, and consequently its accuracy drops to $76.3\%$ (which is approximately the frequency of gold standard zeros in the test set).

Together, these results show that if the defining feature is included in the training data, the black-box model is able to reconstruct the already known definition almost perfectly. However, it lacks the ability to make correct predictions when it is deprived of the defining features during training or testing.
\section{Related Work}

The problem described in our work has not been discussed in the literature so far. 
First, our detection procedure must not be confused with standard feature selection 
\citep{HastieETAL:08}. These techniques are aimed at improving model prediction while our method is a data analysis technique, aimed at distinguishing defining features from \emph{powerful features}, i.e. other features that are highly correlated with the target labels. Such features can also take the form of \emph{shortcuts} where a spurious correlation between an irrelevant input feature and the target label is exploited by the model \citep{GeirhosETAL:20}. This has been shown to happen in natural language processing, image processing, and medical data science \citep{RibeiroETAL:16,LapuschkinETAL:19,DeGraveETAL:21}. 
The discrimination between powerful features and defining features is achieved by step 2 of our method since defining features will result in nullification even of powerful features, while powerful features will not cause this effect. 

Defining features also need to be distinguished from \emph{leaking features}. \cite{RossetETAL:09} and \cite{KaufmanETAL:11} define leakage in data mining as the \say{introduction of information about the data  mining target, which should not be legitimately available to mine from}. Examples for leaking input features are inclusion of the target itself, or of future information in a prediction task.
In contrast to illegitimate data mining from leaking features, it is perfectly legitimate to diagnose an illness according to a consensus definition, and the involved physiological measurements are a legitimate part of input representations in machine learning. However, a validity problem arises by a double usage --- directly as input features and indirectly in the definition of outputs --- in difference to leaking features that are illegitimate per se. 

\section{Discussion}

We described a problem that potentially arises from the common practice of generating machine learning data by automatically labeling clinical datasets according to consensus definitions. Since consensus definitions are indirect measurements based on fundamental clinical measurements, if the latter are included as input features in machine learning datasets, they become defining features causing machine learning models to learn nothing else but to reconstruct the known consensus definition.

We demonstrated the applicability of our procedure to detect three cases of indirect measurements based on three fundamentally different definitions: 
For SOFA$_{liver}$, the definition consists of a single additive component, which naturally extends to definitions based on \textit{sums} of multiple components. For SOFA$_{kidney}$, the definition is based on the more complex \textit{maximum}-operation between several components. In the final case of Sepsis-3, the definition is based on a logical \textsc{and}-operation which can be expressed as a \textit{product} of two values. 
In all three cases, our procedure was able to detect a validity problem caused by labeling via indirect measurements. Most of the consensus definitions we are aware of can be modeled using these three basic operations. 

The experiments conducted in this paper follow a cross-sectional experiment design. However, we believe that our conclusions also extend to a longitudinal design for prognosis where feature measurements at a current point in time are used to predict the disease status at future points in time. In this setting, the problem of indirect target definitions will be introduced by the auto-correlation of the fundamental features in the time series, especially if data imputation methods like last-value carried forward are used in dataset creation.

One way to avoid the described validity problem is to keep the data labeling process and the input features apart. Another option that restitutes the application of consensus-based definitions while keeping all fundamental features in the input representation could be a forecasting setup: In such a scenario, the machine learning goal is to project the time series of clinical measurements into the future, and then to apply the measurements of the consensus definition to predict the sepsis label.



\acks{This research has been conducted in project SCIDATOS (Scientific Computing for Improved Detection and Therapy of Sepsis), funded by the Klaus Tschira Foundation, Germany (Grant number 00.0277.2015). We would like to thank our colleagues at the University Medical Centre Mannheim (UMM), Germany --- Manfred Thiel, Holger Lindner, and Verana Schneider-Lindner --- for permission to use the data generated in SCIDATOS for the experiments presented in this paper.}

\bibliography{ref.bib}

\appendix

\vspace{-.5em}
\section*{Appendix}

\subsection*{Statistical background}

A prerequisite for our detection procedure for defining features is the availability of an expressive and yet intelligible model that can be fitted to data $D = \{(\mathbf{x}^n,{y}^n)\}_{n=1}^N$, where $\mathbf{x}^n$ denotes a $p$-dimensional vector of input features (covariates) and ${y}^n$ a scalar target label.  
As a model we adopt the class of \emph{Generalized Additive Models} (GAMs) that originated in the area of biostatistics \citep{HastieTibshirani:90} to lift the restriction of strictly linear features in generalized linear regression models. GAMs have been used as interpretable white-box models for machine learning \citep{LouETAL:12} and as tools for analyzing black-box models \citep{TanETAL:18}. The key idea of GAMs is decomposing a multivariate function into a sum of usually univariate nonlinear functions $f_k(x_k)$ for each input feature $x_k$, called \emph{feature shapes}, which are learned from the data. The general form of a GAM assumes $Y$ to be a random variable from the exponential family, and $g(\cdot)$ to be a nonlinear but invertible link function: 

\begin{equation}
g(\mathbb{E}[ Y \vert \mathbf{x} ]) = \sum_{k=1}^p f_{k}(x_{k}).
\end{equation}

\noindent{}For example, a Gaussian regression model is obtained by using the identity link function $g(x)=x$, and specifying the distribution of $Y^n$ to be of the zero mean Gaussian subclass of the exponential family:
\begin{align}
Y^n & = \sum_{k=1}^p f_{k}(x_{k}^n) + \epsilon^n, \\
& \textrm{ where } \epsilon^n \sim \mathcal{N}(0,\sigma^2) \textrm{ for } n = 1, \ldots,N. \notag
\end{align}

For our application, we use GAMs that model feature shapes by \emph{regression spline} functions. 
A regression spline represents a non-linear function as a linear combination of so called fixed base functions which are used to transform the input data. 
The estimated weights of this linear combination determine the shape of the non-linear function and thus allow a model with linear structure to learn non-linear functions from data.

In order to define a detection procedure for defining features we need two measures, one that allows us to assess how good a model reproduces its training data, and one that measures its complexity. We use the likelihood-based criterion of \emph{scaled deviance} of a model to assess model fit. \cite{McCullaghNelder:89} define it as a metric proportional to the difference between the log-likelihood $\ell(\mu)$ of a model $\mu$ to the log-likelihood $\ell^\ast$ of the saturated model, which is the model in family that achieves the highest possible likelihood value given the data: 

\begin{equation}
    D^\ast_\mu = 2(\ell^\ast - \ell(\mu)).
\end{equation}

\noindent{}The saturated model corresponds to an exact fit. 
For a Gaussian model with known variance $\sigma^2$ and independent samples, the scaled deviance is:

\begin{equation}
    D^\ast_\mu = 2(\ell^\ast - \ell(\mu)) = \sum_{n=1}^N \frac{(y^n-\mu^n)^2}{\sigma^2}.
\end{equation}

Following \cite{HastieTibshirani:86}, we use the \emph{percentage of deviance explained} to make the metric more interpretable, and denote it by $D^2$:

\begin{equation}
    D^2(\mu) = 1 - \frac{D^\ast_\mu}{D_{\mu_0}}, 
\end{equation}

\noindent{}where $D_{\mu_0}$ is the deviance for the model $\mu_o$  which is the simplest possible model using only a intercept term and no other predictor. This model just predicts a constant, namely the average $\Bar{y}$ of all data points. The likelihood of this model is the lowest possible, thus $D^2(\mu) \in [0,1]$, where $0$ ($0\%$) means that the model $\mu$ fits the data as good as $\mu_0$, and $1$ ($100\%$) means that $\mu$ fits the data as good as the saturated model. 
For a Gaussian GAM with known variance and independent samples, $D^2$ is 

\begin{equation}
    D^2(\mu) = 1 - \frac{\sum_{n=1}^N (y^n-\mu^n)^2}{\sum_{n=1}^N (y^n-\Bar{y})^2}. 
\end{equation}

\noindent{}In this case, $D^2$ equals the coefficient of determination $R^2$. It is thus regarded as a generalization of the latter.

The typical model complexity measure employed for regression models are the \emph{degrees of freedom}, corresponding to the number of model parameters estimated from the data. This measure cannot be directly applied to GAMs due to the involved smoothing process and the penalized regression estimation procedure. Taking into account all the complexities involved with GAMs, \cite{Wood:17} has adapted the concept and  defined a measure called \emph{effective degrees of freedom (edf)} which we use to measure model complexity.

A further concept that is important in our detection procedure for defining features is that of consistency of an estimator. It is one of the most fundamental properties of an estimator and a necessary condition for model interpretability.

\begin{definition}[Consistency]
\hspace{-0.5mm}Let \hspace{-0.3mm}$M := \left\{ p_\theta \colon \theta \in \Theta\right\}$ be a parametric statistical model where $\theta \mapsto p_\theta$ is injective. Further, let $p_{\theta_0} \in M$ denote the true model of the data generating process for a dataset $D = \{(\mathbf{x}^n,{y}^n)\}_{n=1}^N$. Then an estimator ${\theta}_N$ is called \textit{consistent} iff for all $\epsilon > 0$ holds 
    \begin{align*}
 		P\left( \vert {\theta}_N - \theta_0 \vert  > \epsilon \right)  \xrightarrow{N \to \infty} 0.
    \end{align*}

\end{definition}

Consistency of an estimator guarantees that the probability of learning the true data-generating model converges to $1$ as the sample size increases, given that the true model is among the candidates.  Note that consistency holds for spline based GAMs \citep{HastieTibshirani:90,Wood:17,Heckman:86}, but has not been shown for variants where the features shapes are modeled by non-spline machine learning methods like deep neural networks (NAMs) \citep{AgarwalETAL:20}.
The consistency property of GAMs allows us to identify defining features as those that approximate the data generating process with a non-zero feature shape, and features that are not related to the data generation process as those with constant zero feature shapes.

\begin{proposition}[Nullification.]
\label{prop:nullification}
Let $p^{\textrm{GAM}}_{{\theta}_N}$ be a GAM that optimizes the likelihood of data $D = \{(\mathbf{x}^n,{y}^n)\}_{n=1}^N$, $\mathbf{x} = (x_1,\dots,x_u,\dots,x_p)$, that has been produced by a data labeling function $l: (x_1,\dots,x_u) \mapsto y$. Furthermore, assume that $l$ can be approximated by a model in $M^{\textrm{GAM}} = \{ p^{\textrm{GAM}}_{\theta}: \theta \in \Theta \}$. Then any feature $x_{k\leq u}$ used by 
the data labeling function $l$ will have a non-zero feature shape $\hat{f}_k(x_k) \overset{p}{\rightarrow} f_k(x_k)\neq 0$, and any other feature $x_{k > u}$ in the feature set will have a feature shape of a constant zero function $\hat{f}_k(x_k) \overset{p}{\rightarrow} f_k(x_k) = 0$. 
\end{proposition}

\noindent\textbf{\emph{Proof sketch.}} The proposition follows directly from the consistency of maximum likelihood estimators for GAMs. This has been shown, for example, by \cite{Heckman:86} for GAMs based on cubic regression splines. By consistency, the maximum likelihood estimator $\theta_N$ will converge in probability to the data generating parameters $\theta_0$. Since the model $M^{\textrm{GAM}} = \{ p^{\textrm{GAM}}_{\theta}: \theta \in \Theta \}$ is identifiable, as by the injectivity of the mapping $\theta \mapsto p_\theta$, the data generating parameters $\theta_0$ will identify the data generating model $p^{\textrm{GAM}}_{{\theta}_0}$. By the additive structure of this model, only features determining the feature-label relations in the data $D = \{(\mathbf{x}^n,{y}^n)\}_{n=1}^N$ have non-zero feature shapes, and the feature shapes of all other features in the feature set have constant zero values. $\square$

\subsection*{Detection procedure for defining features}

Based on GAMs, the $D^2$ and $edf$ metrics, and the nullification criterion, we define a detection procedure for defining features that is based on the idea that in the presence of defining features, a simple model based solely on such features will suffice to nearly perfectly reproduce the gold standard target. The detection procedure for defining features starts from a set of candidate features, for example, identified by the absolute magnitude of their bivariate correlation with the labels, and then searches for the model with highest deviance and lowest degrees of freedom over a set of candidate models $\mathcal{M_C}$ (step 1), and confirms that the identified features are indeed defining since all other features except the ones found in the first step are nullified (step 2) in an extended model. 

\begin{definition}[Detection procedure]
\label{def:circularity_test}
Given a dataset of feature-label relations $D = \{(\mathbf{x}^n,{y}^n)\}_{n=1}^N$ where $\mathbf{x} = (x_1, x_2, \ldots, x_p)$ is a $p$-dimensional feature vector, let $C \subseteq \mathcal{P}(\{1,\ldots, p\})$ indicate the set of candidate defining features in dataset $D$, and let $\mathcal{M_C} \coloneqq \{ \mu_c \colon c \in C\}$ be the set of models obtained by fitting a GAM including a feature shape for every feature in $c$ (and nothing else) to the data $D$. A set of \textit{defining features}  $c^*$ is detected by applying the following two-step detection procedure:
\begin{enumerate}
    \item $c^* = \operatorname*{argmax}_{c \in C} D^2(\mu_c)$ where $D^2(\mu_{c^*})$ is close to $1$, and in case the maximizer is not unique,  the maximizer is chosen such that the associated GAM $\mu_{c^*}$ has the smallest effective degrees of freedom.
    \item The feature shapes of every feature $x_j : j \in \{1,\ldots,p\} \setminus c^*$ added to the GAM $\mu_{c^*}$ is nullified in the resulting model.
\end{enumerate}
\end{definition}

\end{document}